\documentclass[conference]{IEEEtran}
\IEEEoverridecommandlockouts
\usepackage{cite}
\usepackage{amsmath,amssymb,amsfonts}
\usepackage{algorithmic}
\usepackage{graphicx}
\usepackage{textcomp}
\usepackage{xcolor}
\def\BibTeX{{\rm B\kern-.05em{\sc i\kern-.025em b}\kern-.08em
    T\kern-.1667em\lower.7ex\hbox{E}\kern-.125emX}}
\begin{document}

\title{A Comparison of the Taguchi Method and Evolutionary Optimization in Multivariate Testing}

\author{\IEEEauthorblockN{Jingbo Jiang}
\IEEEauthorblockA{\textit{University of Pennsylvania} \\
Philadelphia, USA \\
yanmoviola@gmail.com}
\and
\IEEEauthorblockN{Diego Legrand}
\IEEEauthorblockA{\textit{Criteo} \\
Paris, France \\
legrand.diego@gmail.com}
\and
\IEEEauthorblockN{Robert Severn}
\IEEEauthorblockA{\textit{Evolv Technologies} \\
San Francisco, USA \\
robert.severn@evolv.ai}
\and
\IEEEauthorblockN{Risto Miikkulainen}
\IEEEauthorblockA{\textit{The University of Texas at Austin} \\
  \textit{and Cognizant Technology Solutions}\\
Austin and San Francisco, USA \\
risto@cs.utexas.edu,risto@cognizant.com}
}

\maketitle

\begin{abstract}
Multivariate testing has recently emerged as a promising technique in web interface design. In contrast to the standard A/B testing, multivariate approach aims at evaluating a large number of values in a few key variables systematically. The Taguchi method is a practical implementation of this idea, focusing on orthogonal combinations of values. It is the current state of the art in applications such as Adobe Target. This paper evaluates an alternative method: population-based search, i.e. evolutionary optimization. Its performance is compared to that of the Taguchi method in several simulated conditions, including an orthogonal one designed to favor the Taguchi method, and two realistic conditions with dependences between variables. Evolutionary optimization is found to perform significantly better especially in the realistic conditions, suggesting that it forms a good approach for web interface design and other related applications in the future.
\end{abstract}


\section{Introduction}
In e-commerce, designing web interfaces (i.e.\ web pages and
interactions) that convert as many users as possible from casual
browsers to paying customers is an important goal
\cite{ash:book12,salehd:book11}. While there are some well-known
design principles, including simplicity and consistency, there are
often also unexpected interactions between elements of the page that
determine how well it converts. The same element, such as a headline,
image, or testimonial, may work well in one context but not in
others---it is often hard to predict the result, and even harder to
decide how to improve a given page.

An entire subfield of information technology has emerged in this area,
called conversion rate optimization, or conversion science. The
standard method is A/B testing, i.e.\ designing two different versions
of the same page, showing them to different users, and collecting
statistics on how well they each convert
\cite{kohavi:encyclopedia16,webdesign,taguchifield,conversion,testing}.  This process allows
incorporating human knowledge about the domain and conversion
optimization into the design, and then testing their effect. After
observing the results, new designs can be compared and gradually
improved. The A/B testing process is difficult and time-consuming:
Only a very small fraction of page designs can be tested in this way,
and subtle interactions in the design are likely to go unnoticed and
unutilized.

An alternative to A/B is multivariate testing, where all value
combinations of a few elements are tested at once
\cite{kohavi:encyclopedia16,conversion,ab}.  While this process captures
interactions between these elements, only a very small number of
elements is usually included (e.g.\ 2-3); the rest of the design space
remains unexplored. The Taguchi method \cite{rao:biotech08,testing} is a
practical implementation of multivariate testing. It avoids the
computational complexity of full multivariate testing by evaluating
only orthogonal combinations of element values. Taguchi is the current
state of the art in this area, included in commercial applications
such as the Adobe Target \cite{Adobe}. However, it assumes that the
effect of each element is independent of the others, which is unlikely
to be true in web interface design. It may therefore miss interactions
that have a significant effect on conversion rate.

This paper evaluates an alternative approach based on evolutionary
optimization. Because such search is based on intelligent sampling of
the entire space, instead of statistical modeling, it can potentially
overcome those shortcomings. Crossover and mutation can traverse
massive solution spaces efficiently, discovering dependencies and
using them as building blocks \cite{goldberg,nn,deb,hormoz}. It can therefore search
the space in a more comprehensive manner, and effectively find
solutions that the other methods miss.

\begin{figure}[!t]
  \begin{center}
    \includegraphics[width=\columnwidth]{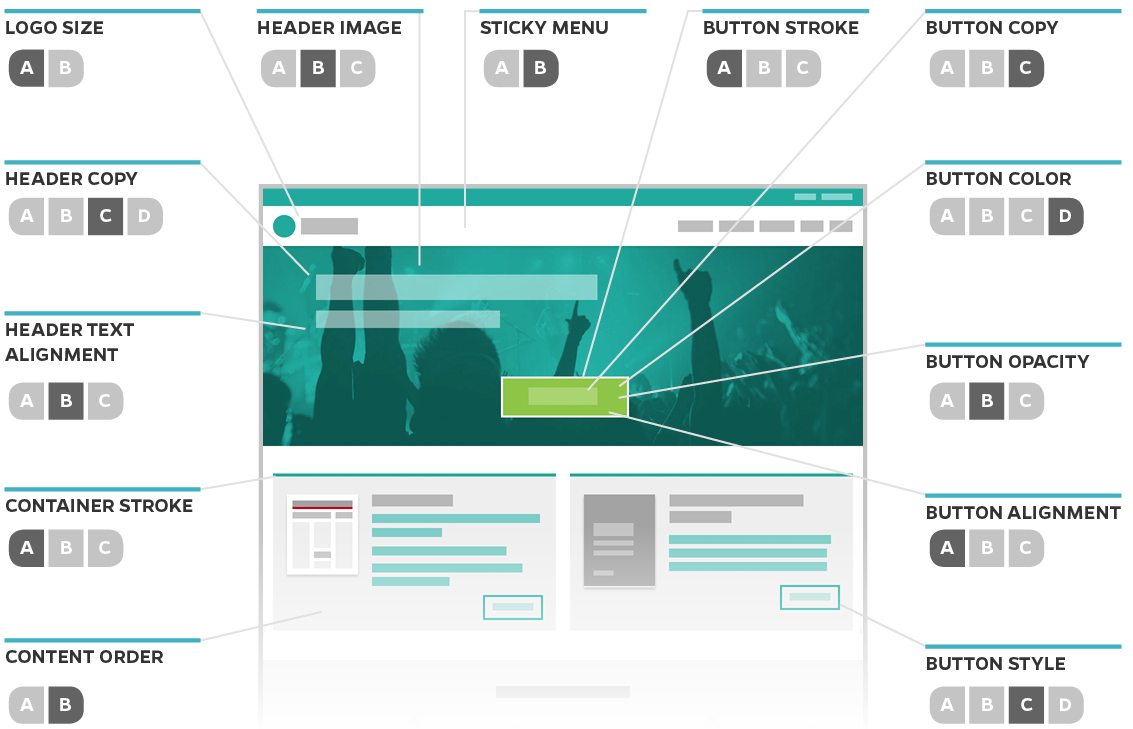}
    \caption{Web Interface Design as an Optimization Problem. In this
      example, 13 elements of the page each have 2-4 possible values,
      resulting in 1.1M combinations. The goal is to find combinations
      that make it as likely as possible for the user to click on the
      action button in the middle. The state-of-the-art multivariate
      testing method, Taguchi, evaluates only orthogonal combinations
      and therefore misses nonlinear interactions between elements. In
      contrast, population-based search in Ascend is sensitive to such
      interactions; it can therefore find good solutions that the
      Taguchi method misses.}
    \label{fg:combinations}
  \end{center}
\end{figure}

This idea is implemented in Ascend by Evolv web-interface optimization
system, deployed in numerous e-commerce websites of paying customers
since September 2016 \cite{evo,iaai}. Ascend uses a customer-designed search
space as a starting point (Figure~\ref{fg:combinations}).  It consists
of a list of elements on the web page that can be changed, and their
possible alternative values, such as a header text, font, and color,
background image, testimonial text, and content order. Ascend then
automatically generates web-page candidates to be tested, and improves
those candidates through evolutionary optimization.

This paper compares the evolutionary approach in Ascend to the
state-of-the-art statistical method of Taguchi optimization in
simulated experiments. The results show that evolution is indeed more
powerful optimizer, especially under realistic conditions where there
are nonlinear dependencies between variables. It is therefore a
promising foundation for designing optimization applications in the
future, and increases the potential for more powerful AI applications
in related fields.

\section{The Taguchi Method}
Ideally, the best web interface design would be decided based on full factorial multivariate testing. That is, each possible combination of $N$ variables with $K$ values would be implemented as a candidate. For example, a variable might be the color or the position of a button; the values would then be the possible colors and positions. A full factorial analysis would require testing all $K^N$ combinations, which is prohibitive in most cases.

 Instead, the Taguchi method specifies a small subset of these combinations to test using orthogonal arrays. An Taguchi orthogonal array is a matrix where each column corresponds to a variable and each row to a candidate to test. Each value represents the setting for a given variable and experiment.  It has the following properties: 
 
\begin{itemize}
\item The dot product between any two normalized column  vectors is zero. 
\item For every variable column, each value appears the same amount of times.
\end{itemize}
 
There are multiple ways of creating orthogonal arrays \cite{orth,orth2}. Table \ref{taguchieg} shows an example of an orthogonal array of nine combinations, resulting from testing four variables of three values each.\\

\begin{table}[ht]
\centering
\begin{tabular}{c|ccccc}
\hline
Candidate&Var 1&Var 2&Var 3&Var 4&Performance\\
\hline
1 &0&0&0&0&p1\\
2 &0&1&2&1&p2\\
3 &0&2&1&2&p3\\
4 &1&0&2&2&p4\\
5 &1&1&1&0&p5\\
6 &1&2&0&1&p6\\
7 &2&0&1&1&p7\\
8 &2&1&0&2&p8\\
9 &2&2&2&0&p9\\
\hline
\vspace*{2ex}
\end{tabular}
\caption{Example Taguchi array of four variables with three values each\label{taguchieg}}
\end{table}

To compute the effect of a specific variable value, we average the performance scores of the candidates corresponding to combinations for that value setting. Because in an orthogonal array, all values of the other variables are tested an equal amount of times, their effects cancel out, assuming each variable is independent\cite{orth2}. For example, to compute the effect of value 2 of variable 3 in table\ref{taguchieg}, we average the scores of candidates 2, 4 and 9. Similarly, for value 1, we average the scores of candidates 3, 5 and 7.  \\

In a Taguchi experiment, all the candidates (rows) in the orthogonal table are tested, and the scores for candidates that share the same value for each variable are averaged in this manner. We can then predict a best performing combination by selecting, for each variable, the value with the best such average score. 

The Taguchi method is a practical approximation of factorial testing. However, the averaging steps assume that the effects of each variable are independent, which may or may not hold in real-world experiments. In contrast, population-based search makes no such assumptions, as will be discussed next.

\section{Evolutionary Optimization}
\begin{table*}[!t]
  \begin{minipage}{\columnwidth}
    \begin{center}
      \begin{tabular}{l|l}
\hline
$CR$ &true conversion rate\\
\hline
$c$ &candidate\\
\hline
$n$& the number of variables\\
\hline
$W^0$& bias (i.e. CR of the control candidate)\\
\hline
$W^1_i(c)$& impact of the candidate $c$'s value for variable $i$\\
\hline
$W^2_{jk}(c)$& interaction between candidate $c$'s values of \\
&variables $j$ and $k$\\
\hline
\vspace*{2ex}
\end{tabular}
\caption{Evaluator Denotation\label{evaluator}}
    \end{center}
  \end{minipage}
  \hfill
  \begin{minipage}{\columnwidth}
    \begin{center}
      \begin{tabular}{l|l}
\hline
Parameter & Value \\\hline
Number of generations & 8 \\
Mutation weight & 0.01\\
Elite percentage &20\%\\
Evaluator bias (i.e. control's conversion rate) $W^0$& 0.05\\\hline
\vspace*{2ex}
\end{tabular}
\caption{\label{tab:widgets}Default parameter setting}
\vspace*{0.35in}
\label{table1}
    \end{center}
  \end{minipage}
\end{table*}
Evolutionary optimization is a broadly used method for combinatorial problems, building on population-based search. It has certain advantages compared to diagnostic methods, including fewer restrictions on input variables, robustness to environmental changes, good scale-up to large and high-dimensional spaces, and robustness to deceptive search spaces and nonlinear interactions \cite{evogood}.  The main idea is that instead of constructing the winning combination though independence assumptions (as in Taguchi), the winner is searched for using crossover and mutation operators. 

The evolution algorithm used in this paper is that of Ascend by Evolv, a conversion optimization product for web interfaces \cite{evo}. The basic unit of the method is the candidate's genome, which is a list representing the elements and values of the web interface. For example, the genome $\textbf{[2,4,5,3]}$ defines a web page with four changeable parts, i.e. genes, with 2, 4, 5, and 3 different choices each. The choices are represented as one-hot vectors, and are concatenated to form the genome. The control candidate is the genome representing default web settings, consisting of the first dimension in each vector: $$[[1, 0], [1, 0, 0, 0], [1, 0, 0, 0, 0], [1, 0, 0]]$$The candidates in the first generation consist of all genomes that are one gene different from the control. Thus, the number of candidates in this generation is the sum, over all genes, of the number of values minus one, i.e. $1+3+4+2=10$ in this example. In all future generations, the total number of candidates stays the same; a certain percentage of candidates (e.g. 20\%) are chosen as elites, staying on to the next generation. The remaining (e.g. 80\%) candidates are  formed by crossover from those elites \cite{evoselect}.  The encoding of each candidate also has a chance to mutate (i.e. specify a different choice for each part), according to probabilities specified in the parameter setting. The evolution process ends after a prespecified number of generations or after a suitable candidate is found.

\begin{figure}[!t]
  \begin{center}
    \includegraphics[width=\columnwidth]{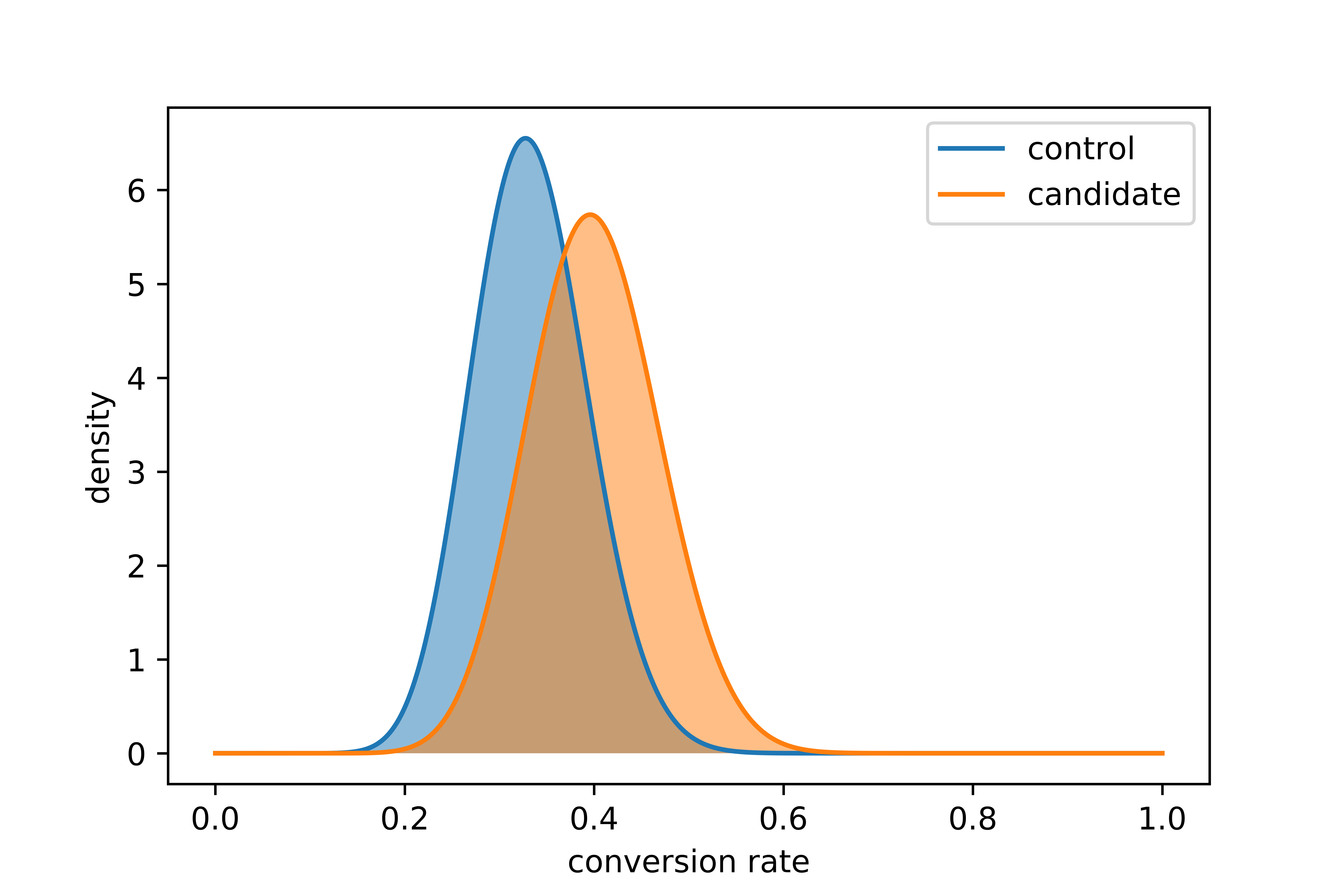}
      \vspace{-0.6cm}
    \caption{Probability distribution of control and target candidate conversion rates. The area under the candidate curve that is above the control curve stands for the probability to beat control. Using that probability as the measure of performance, instead of the estimated conversion rate, leads to more reliable results.}
    \label{pbc}
  \end{center}
\end{figure}

In this paper, the key value of evaluating the performance from different approaches is conversion rate, which is the ratio of people that convert to the total visitor of the web page. After an evolutionary simulation, a prior estimate of the conversion rate is obtained as the average of all candidates tested. A probability to beat control is computed for each candidate based on this prior and its individual estimate. To compute the probability to beat control for a target candidate, a probability distribution of conversion rate is built for the control and the candidate, based on data collected from experiment, as demonstrated in Figure \ref{pbc}. Then the proportion of area under the curve of candidate conversion rate distribution where it beats the one of control is computed as the probability to beat control for the candidate, and the one with the highest probability is selected as the winner.

\section{Evaluation in a Simulator}

The performance of the Taguchi method and Evolutionary optimization was measured in simulated experiments of designing web interfaces for maximum conversion rate. In the simulation, an evaluator is first constructed to calculate a candidate's true conversion rate based on the values it specifies for each variable. Simulated traffic is then distributed to candidates and conversions are assigned probabilistically based on candidates' true conversion rate. The observed conversion rates are then used as the scores of the candidates in Taguchi and evolution methods.

By setting the parameters in Table \ref{evaluator}, different kinds of evaluators can be defined.  The conversion rate of simple linear evaluator is based on only bias and weight for each individual variable: 
\begin{equation}\label{linear}
CR[c] = W^0+\sum_{i=1}^nW_i^1(c).
\end{equation}The bias represents the conversion rate of the control candidate; the different choices for each variable add or subtract from the control rate. A non-linear evaluator is designed to include interactions between variables:
\begin{equation}\label{nonlinear}
CR[c] = W^0+\sum_{i=1}^nW_i^1(c)+\sum_{j=1}^n\sum_{k=j+1}^nW_{j,k}^2(c).
\end{equation}In addition to bias and individual variable contributions, it includes contributions for each pair of variables.

Both the Taguchi candidates and the evolution candidates are represented in the same way, as concatenations of one-hot vectors representing the values for each variable in the Taguchi method, and actions for each gene in evolution. The total traffic for the Taguchi method and evolution algorithm is set to be equal, distributed evenly to all Taguchi candidates, but differently for evolution candidates based on how many generations they survive.

Table \ref{table1} specifies the parameter settings used in all experiments with evolution, and the evaluator bias rate.

\section{Experimental Results}

Three experiments were run comparing the Taguchi method with evolutionary optimization: two experiments where the variables had independent effect, one with a uniform and the other with varied number of  values;  and one experiment with dependencies between pairs of variables. In addition to comparing the ability of these methods to find good candidates as the final result of the experiment, their performance during the experiment was also compared. The result curves demonstrate statistical means and 95\% credible intervals of 20 repeated experiments, under the same example settings in each section.
\begin{figure*}[!t]
  \begin{minipage}{\columnwidth}
    \begin{center}
      \includegraphics[width=\textwidth]{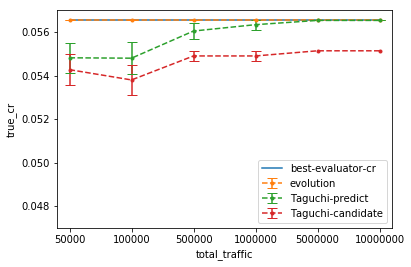}
	\caption{\label{fig:51}\textbf{True conversion rate performance on the [3,3,3,3] setting for the Taguchi method and evolution with increasing amount of traffic.  The ``evolution'' line is the best candidate chosen by evolution algorithm; the ``Taguchi-predict'' line is the combined candidate from Taguchi variable analysis; ``Taguchi-candidate'' is the highest scored candidate in original input Taguchi array. Evolution algorithm performs significantly better with small amount of traffic; after about 500,000, both methods perform similarly.}}
    \end{center}
  \end{minipage}
  \hfill
  \begin{minipage}{\columnwidth}
    \begin{center}
      \includegraphics[width=\textwidth]{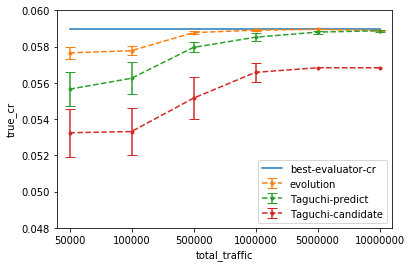}
\caption{\label{fig:52}\textbf{True conversion rate performance with a more complex genome. Both methods take longer to find good candidates; Taguchi is now comparable to evolution only with the highest amounts of traffic.}}
    \vspace*{0.7in}
    \end{center}
  \end{minipage}
\end{figure*}
\subsection{Independent Variables with Uniform Values}

The Taguchi method assumes that the variables are independent. The first experiment was designed accordingly: It uses a linear evaluator that assumes all changes are independent, and a simple genome that results in few rows  in the Taguchi array.  These are the ideal conditions for the Taguchi method, and it is expected to perform well.
The best settings for the Taguchi method are those with uniform numbers of values across all variables \cite{Adobe}: 

\begin{itemize}
    \item 
Setting 1: 

Three variables with two values each, i.e. $\textbf{[2, 2, 2]}$,  with $2^3=8$ combinations, resulting in four rows;
\item

Setting 2: 

Four variables with three values each, i.e. $\textbf{[3, 3, 3, 3]}$, with $3^4=81$  combinations, resulting in nine rows; and
\item
Setting 3: 

Five variables with four values each, i.e. $\textbf{[4, 4, 4, 4, 4]}$, with $4^5=1024$  combinations, resulting in 16 rows.
\end{itemize}

The Taguchi arrays for these settings can be found in orthogonal array libraries \cite{lib}. The learning curves under all three settings are similar, so Setting 2 will be used as an example.

Figure \ref{fig:51} shows the true conversion rates of the best candidates under Setting 2 with increasing traffic. The true conversion rate for the best evolution candidate is steady and high at all traffic values. The best predicted Taguchi candidate's true conversion rate lags behind evolution with low traffic, but eventually catches up as traffic increases. The best tested Taguchi candidate remains significantly below both curves, which shows that Taguchi method does achieve an improvement from its original tested candidates to the predicted candidates. Thus, under ideal conditions for Taguchi, both methods find equally good solutions given enough traffic (i.e. more than 500,000). With low traffic, the best evolutionary approach performs significantly better.

Since the total traffic is equal with both approaches, and Taguchi method defines a set of candidates to distributed those traffic, it then has a fixed experiment plan that is indifferent to the allocation of traffic. However, evolution algorithm consecutively generates better performed candidates to use the limited traffic through generations. This way, 
it better exploits data to drive selection of good candidates in the end, because it allows testing to focus on good combinations adaptively, especially when total traffic is not sufficient. 

\subsection{Independent Variables with Variable Values}
\label{sc:indep}

In real world applications, such as optimization of commercial websites, the design space may be rather complex; in particular, the number of values for each variable is not likely to be the same.  In the second experiment, while still maintaining independence between variables, the genome structure is changed to:$$\textbf{[3, 6, 2, 3, 6, 2, 2, 6]},$$
i.e. three variables with two value each, two variables with three value each, and three variables with six value each. In this setting with 15,552 combinations, the Taguchi array needs 36 rows \cite{lib}.

\begin{figure*}[!t]
  \begin{minipage}{\columnwidth}
    \begin{center}
      \includegraphics[width=\textwidth]{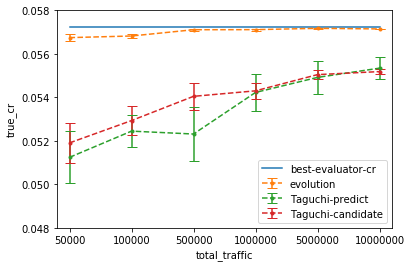}
\caption{\label{fig:53}\textbf{\textbf{True conversion rate
      performance  with interacting variables. Evolutionary
      optimization now results in significantly better candidates at
      all traffic values. The ``Taguchi-predict'' result is
      similar to ``Taguchi-candidate'', suggesting that the interactions render the construction process ineffective.}}}
    \end{center}
  \end{minipage}
  \hfill
  \begin{minipage}{\columnwidth}
    \begin{center}
      \includegraphics[width=\textwidth]{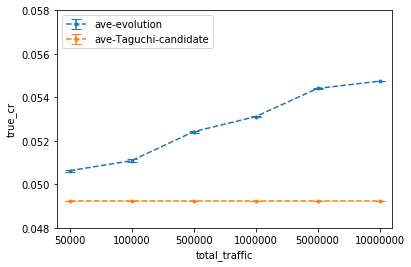}
\caption{\label{fig:54}\textbf{\textbf{Average true conversion rate of candidates with evolution and Taguchi methods during the experiment. While Taguchi candidates do not change, evolution continuously comes up with better candidates, thus increasing performance during the experiment. It therefore forms a good approach for campaigns with fixed duration as well.}}}
    \end{center}
  \end{minipage}
\end{figure*}
The result in Figure \ref{fig:52} shows that with a more complex problem, both evolution and Taguchi require more traffic in order to find good solutions. However, evolution produces significantly better candidates than Taguchi at almost all traffic values: The two methods are comparable only for very high traffic, i.e. greater than 5,000,000. The prediction process of Taguchi still provides a major advantage beyond its input set. 

Similarly to the discussion above, evolution algorithms maintains better ability for choosing good candidates because its effort in distributing data to better candidates reactively. In this more complex scenario, Taguchi method requires more candidates for testing, which greatly decrease the exposure to data for all its candidates, since they are equally tested and distributed with same amount of traffic. This results in a much higher total traffic for Taguchi method to perform similarly as evolution algorithm, which still exploits data on good candidates.

\subsection{Interactions Between Variables}

Another important challenge in real-world applications is that the variables are not likely to be independent. For example, text color and background color may interact---for instance, blue text on a blue background would perform poorly compared to blue text on a white background. The nonlinear evaluator is designed to test the ability of the two methods to handle this kind of interactions. The example uses genome in Section~\ref{sc:indep}.

Figure \ref{fig:53} shows that when the independence assumption for Taguchi method is broken, the best predicted Taguchi candidate's true conversion rate is no longer comparable with evolution's. Furthermore, its predicted best candidate does not even significantly outperform its best tested candidate. Interestingly, the performance of the evolutionary algorithm is not significanly worse with interacting vs. independent variables, demonstrating its ability to adapt to complicated real-world circumstances.

\subsection{Performance During Experiment}

The main goal in conversion optimization is to find good candidates that can be deployed after the experiment. However, in many cases it is also important to not decrease the site's performance much during the experiment. Evolution continuously creates improved candidates as it learns more about the system, whereas the Taguchi method generates a single set of candidates for the entire test---it therefore provides continual improvement on the site even during the experiment.

This principle is illustrated graphically in Figure \ref{fig:54}, using the linear evaluator and genome from Section~\ref{sc:indep} as the example setting. The Taguchi's candidates' average performance stays the same throughout the increasing traffic, whereas evolution's candidates perform, on average, better with more traffic, i.e. while the experiment progresses. It therefore forms a good approach in domains where performance matters during the experiment, in particular in campaigns that run only for a limited duration.

\section{Discussion and Future Work}

The two methods tested in the experiments of this paper, Taguchi and evolution, are both beneficial in web interface design. Taguchi's appeal is its high reduction of rows compared to full factorial combinations, which works best when the genome structure is rather simple. When the genome becomes larger and more complex, its performance falls behind that of evolution. Most importantly, if there are nonlinear interactions between the variables, the method cannot keep up with them: the best-candidate construction does not improve upon its initial best candidates, and it performs much worse than evolution at all traffic values.

In contrast, the process of searching for good candidates in evolution is based on crossover and mutation, and therefore is not affected much by interactions. Evolution discovers good combinations, and constructs future candidates using them as building blocks. As long as the interactions occur within the building blocks, they will be included and utilized the same way as independent contributions. Given how common interactions are in real-world problems, this ability should turn out important in applying optimization to web interface design in the future, enabling more powerful AI applications in related areas.

\section{Conclusion}

This paper demonstrates that (1) even with ideal conditions, the Taguchi method does not exceed performance of evolution, and (2) with low traffic in ideal conditions, evolution performs significantly better. (3)  As the experiment configuration becomes more complex, Taguchi requires more traffic to match evolution's performance. (4) With nonlinear interactions between variables, Taguchi's construction process breaks down, and it no longer improves upon best initial candidates.  (5) In contrast, evolution is able to find good candidates even with nonlinear interactions. Furthermore, (6) evolution improves during the duration of the experiment, making it a good choice for campaigns as well. Evolutionary optimization is thus a superior
technique for improving conversion rates in web interface design.

\bibliographystyle{IEEEtranS}
\bibliography{paper}

\begin{thebibliography}{10}
\providecommand{\url}[1]{#1}
\csname url@samestyle\endcsname
\providecommand{\newblock}{\relax}
\providecommand{\bibinfo}[2]{#2}
\providecommand{\BIBentrySTDinterwordspacing}{\spaceskip=0pt\relax}
\providecommand{\BIBentryALTinterwordstretchfactor}{4}
\providecommand{\BIBentryALTinterwordspacing}{\spaceskip=\fontdimen2\font plus
\BIBentryALTinterwordstretchfactor\fontdimen3\font minus
  \fontdimen4\font\relax}
\providecommand{\BIBforeignlanguage}[2]{{%
\expandafter\ifx\csname l@#1\endcsname\relax
\typeout{** WARNING: IEEEtranS.bst: No hyphenation pattern has been}%
\typeout{** loaded for the language `#1'. Using the pattern for}%
\typeout{** the default language instead.}%
\else
\language=\csname l@#1\endcsname
\fi
#2}}
\providecommand{\BIBdecl}{\relax}
\BIBdecl

\bibitem{Adobe}
Adobe, \emph{Multivariate Test}, 2020, docs.adobe.com/content/help/en/target/
  using/activities/multivariate-test/multivariate-testing.html, retrieved on
  1/2/2020.

\bibitem{ash:book12}
T.~Ash, R.~Page, and M.~Ginty, \emph{Landing Page Optimization: The Definitie
  Guide to Testing and Tuning for Conversions}, 2nd~ed.\hskip 1em plus 0.5em
  minus 0.4em\relax Hoboken, NJ: Wiley, 2012.

\bibitem{evogood}
J.~Branke, \emph{Evolutionary optimization in dynamic environments}.\hskip 1em
  plus 0.5em minus 0.4em\relax Springer Science \& Business Media, 2012, vol.
  Vol. 3.

\bibitem{orth}
A.~E. Brouwer, A.~M. Cohen, and M.~V. Nguyen, ``Orthogonal arrays of strength 3
  and small run sizes,'' \emph{Journal of Statistical Planning and Inference},
  vol. 136, no.~9, pp. 3268--3280, 2006.

\bibitem{deb}
K.~Deb and C.~Myburgh, ``Breaking the billion-variable barrier in real-world
  optimization using a customized evolutionary algorithm,'' \emph{In
  Proceedings of the Genetic and Evolutionary Computation Conference}, pp. pp.
  653--660, 2016, aCM.

\newpage
\bibitem{ab}
E.~Dixon, E.~Enos, and S.~Brodmerkle, ``A/{B} testing of a webpage,''
  \emph{Journal of Advertising Research}, 2011, washington, DC: U.S. Patent and
  Trademark Office.

\bibitem{webdesign}
X.~Dreze and F.~Zufryden, ``Testing web site design and promotional content,''
  \emph{Journal of Advertising Research}, vol.~37, no.~2, pp. 77--91, 1997.

\bibitem{nn}
D.~Floreano, P.~Dürr, and C.~Mattiussi, ``Neuroevolution: from architectures
  to learning,'' \emph{Evolutionary Intelligence}, vol.~1, no.~1, pp. 47--62,
  2008.

\bibitem{goldberg}
D.~E. Goldberg, \emph{Genetic algorithms}.\hskip 1em plus 0.5em minus
  0.4em\relax Pearson Education India, 2006.

\bibitem{orth2}
A.~S. Hedayat, N.~J.~A. Sloane, and J.~Stufken, \emph{Orthogonal arrays: theory
  and applications}.\hskip 1em plus 0.5em minus 0.4em\relax Springer Science \&
  Business Media, 2018.

\bibitem{kohavi:encyclopedia16}
R.~Kohavi and R.~Longbotham, ``Online controlled experiments and {A/B} tests,''
  in \emph{Encyclopedia of Machine Learning and Data Mining}, C.~Sammut and
  G.~I. Webb, Eds.\hskip 1em plus 0.5em minus 0.4em\relax New York: Springer,
  2016.

\bibitem{testing}
R.~Kohavi and S.~Thomke, ``The surprising power of online experiments,''
  \emph{Harvard Business Review}, vol.~95, no.~5, p. pp. 74–82, 2017.

\bibitem{lib}
W.~F. Kuhfeld, 2018, Orthogonal Arrays.
  support.sas.com/techsup/\\technote/ts723.html, retrieved 1/2/2020.

\bibitem{evoselect}
H.~S. N.~D. Miikkulainen, Risto and P.~Long, ``How to select a winner in
  evolutionary optimization?'' \emph{In Computational Intelligence (SSCI), 2017
  IEEE Symposium Series on,}, pp. pp. 1--6, 2017, iEEE.

\bibitem{evo}
R.~Miikkulainen, N.~Iscoe, A.~Shagrin, R.~Cordell, S.~Nazari, C.~Schoolland,
  M.~Brundage, J.~Epstein, R.~Dean, and G.~Lamba, ``Conversion rate
  optimization through evolutionary computation,'' \emph{In Proceedings of the
  Genetic and Evolutionary Computation Conference}, pp. pp. 1193--1199, 2017,
  ACM.

\bibitem{iaai}
R.~Miikkulainen, N.~Iscoe, A.~Shagrin, R.~Rapp, S.~Nazari, P.~McGrath, and
  C.~S. et~al, ``Sentient {A}scend: {AI}-based massively multivariate
  conversion rate optimization,'' \emph{In Proceedings of the Thirtieth
  Innovative Applications of Artificial Intelligence Conference. AAAI}, 2018.

\bibitem{conversion}
W.~W. Moe and P.~S. Fader, ``Dynamic conversion behavior at e-commerce sites,''
  \emph{Management Science}, vol.~50, no.~3, pp. 326--335, 2004.

\bibitem{rao:biotech08}
R.~S. Rao, C.~G. Kumar, R.~S. Prakasham, and P.~J. Hobbs, ``The taguchi
  methodology as a statistical tool for biotechnological applications: A
  critical appraisal,'' \emph{Biotechnology Journal}, vol.~3, pp. 510--523,
  2008.

\bibitem{salehd:book11}
K.~Salehd and A.~Shukairy, \emph{Conversion Optimization: {T}he Art and Science
  of Converting Prospects to Customers}.\hskip 1em plus 0.5em minus 0.4em\relax
  Sebastopol, CA: O'Reilly Media, Inc., 2011.

\bibitem{hormoz}
H.~Shahrzad, D.~Fink, and R.~Miikkulainen, ``Enhanced optimization with
  composite objectives and novelty selection,'' \emph{In Proceedings of the
  2018 Conference on Artificial Life}, 2018.

\bibitem{taguchifield}
G.~Taguchi and J.~Rajesh, ``New trends in multivariate diagnosis,''
  \emph{Sankhyā: The Indian Journal of Statistics, Series B}, pp. 233--248,
  2000.

\end{thebibliography}

\end{document}